\documentclass[conference]{IEEEtran}
\IEEEoverridecommandlockouts

\usepackage{amsmath,amsfonts,bm}

\def\eqref#1{equation~\ref{#1}}

\def\1{\bm{1}}

\def\mC{{\bm{C}}}

\def\mF{{\bm{F}}}

\def\mP{{\bm{P}}}

\def\mR{{\bm{R}}}
\def\mS{{\bm{S}}}
\def\mT{{\bm{T}}}

\DeclareMathAlphabet{\mathsfit}{\encodingdefault}{\sfdefault}{m}{sl}
\SetMathAlphabet{\mathsfit}{bold}{\encodingdefault}{\sfdefault}{bx}{n}

\def\gC{{\mathcal{C}}}

\def\gR{{\mathcal{R}}}
\def\gS{{\mathcal{S}}}
\def\gT{{\mathcal{T}}}

\def\gV{{\mathcal{V}}}
\def\gW{{\mathcal{W}}}

\usepackage{cite}
\usepackage{hyperref} 
\usepackage{url}

\usepackage{amsmath,amssymb,amsfonts}
\usepackage[ruled,vlined,noend]{algorithm2e}
\usepackage{graphicx}
\usepackage{bbm}
\usepackage{textcomp}
\usepackage{wrapfig}
\usepackage{xcolor}
\usepackage{booktabs}      
\usepackage{amsfonts}     
\usepackage{nicefrac}      
\usepackage{microtype} 
\usepackage{enumitem}
\usepackage{multirow,tabularx}
\usepackage{amsmath}
\usepackage{graphicx}
\usepackage{amssymb,amsthm}
\usepackage{inconsolata}
\usepackage{epsfig}
\usepackage{makecell}
\usepackage{subfigure}
\theoremstyle{definition}

\usepackage{bm}
\usepackage[numbers,sort&compress]{natbib}

\def\BibTeX{{\rm B\kern-.05em{\sc i\kern-.025em b}\kern-.08em
    T\kern-.1667em\lower.7ex\hbox{E}\kern-.125emX}}
\begin{document}

\title{Knapsack Optimization-based Schema Linking for \\ LLM-based Text-to-SQL Generation}
\author{
Zheng Yuan\textsuperscript{1}, Hao Chen\textsuperscript{2 *}\thanks{* Corresponding author}, Zijin Hong\textsuperscript{1}, Qinggang Zhang\textsuperscript{1} \\ Feiran Huang\textsuperscript{3}, Qing Li\textsuperscript{1},~\IEEEmembership{Fellow,~IEEE}, and Xiao Huang\textsuperscript{1}\\ 
\textsuperscript{1}The Hong Kong Polytechnic University\\
\textsuperscript{2}City University of Macau, \textsuperscript{3}Jinan University \\
{\{yzheng.yuan, zijin.hong, qinggangg.zhang\}@connect.polyu.hk} \\
{sundaychenhao@gmail.com}; {huangfr@jnu.edu.cn}; \{qing-prof.li, xiao.huang\}@polyu.edu.hk
}
\maketitle

\begin{abstract}
Generating SQLs from user queries is a long-standing challenge, where the accuracy of initial schema linking significantly impacts subsequent SQL generation performance.  However, current schema linking models still struggle with missing relevant schema elements or an excess of redundant ones. A crucial reason for this is that commonly used metrics, recall and precision, fail to capture relevant element missing and thus cannot reflect actual schema linking performance. Motivated by this, we propose enhanced schema linking metrics by introducing a \textbf{restricted missing indicator}. Accordingly, we introduce \textbf{\underline{K}n\underline{a}psack optimization-based \underline{S}chema \underline{L}inking \underline{A}pproach (KaSLA)}, a plug-in schema linking method designed to prevent the missing of relevant schema elements while minimizing the inclusion of redundant ones. KaSLA employs a hierarchical linking strategy that first identifies the optimal table linking and subsequently links columns within the selected table to reduce linking candidate space. In each linking process, it utilizes a knapsack optimization approach to link potentially relevant elements while accounting for a limited tolerance of potentially redundant ones. With this optimization, KaSLA-1.6B achieves superior schema linking results compared to large-scale LLMs, including DeepSeek-V3 with the state-of-the-art (SOTA) schema linking method. Extensive experiments on Spider and BIRD benchmarks verify that KaSLA can significantly improve the SQL generation performance of SOTA Text2SQL models by substituting their schema linking processes. The code is available at \url{https://github.com/DEEP-PolyU/KaSLA}.
\end{abstract}


\section{Introduction}
\label{sec:introduction}
\begin{figure}[!t]
	\centering
        \includegraphics[width=1.0\linewidth]{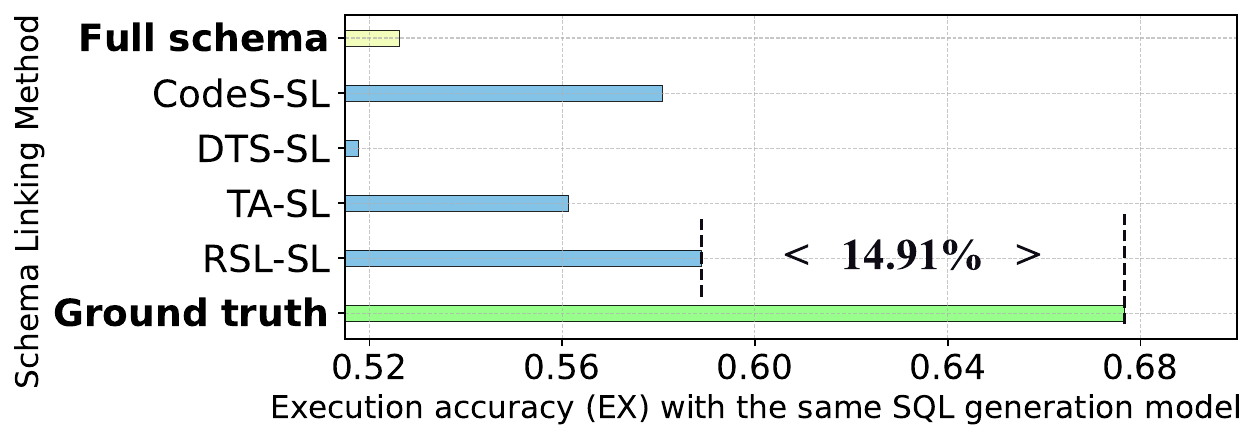}
        \caption{Performance comparison of different schema linking methods on the BIRD-dev dataset,  using CodeS-15B as the SQL generation model.} 
        \label{fig:intro-figure_text-to-SQL}
\end{figure}

With the advent of large language models (LLMs)~\cite{achiam2023gpt,dubey2024llama3}, LLM-based text-to-SQL is emerging as the next-generation database interface for unskilled users~\cite{hong2024next}. Typically, text-to-SQL frameworks employ a two-stage process: first linking user queries to database schema elements (tables and columns), then generating the corresponding SQL statements. Accurate SQL generation relies on correctly identifying these schema elements, since errors in this initial linking phase inevitably propagate to subsequent SQL construction. Inaccurate schema linking, such as overlooking relevant schema elements or linking to irrelevant ones, consistently produces erroneous SQL. This challenge is shown in Fig~\ref{fig:intro-figure_text-to-SQL}. When using CodeS-15B, a pre-trained LLM for SQL generation, on the BIRD-dev benchmark, there is a significant 14.91\% performance gap between state-of-the-art schema linking methods and ground-truth linking results.
Consequently, enhancing schema linking accuracy represents a critical research frontier with significant potential to advance text-to-SQL capabilities.

\begin{figure}[!t]
	\centering
        \includegraphics[width=1.0\linewidth]{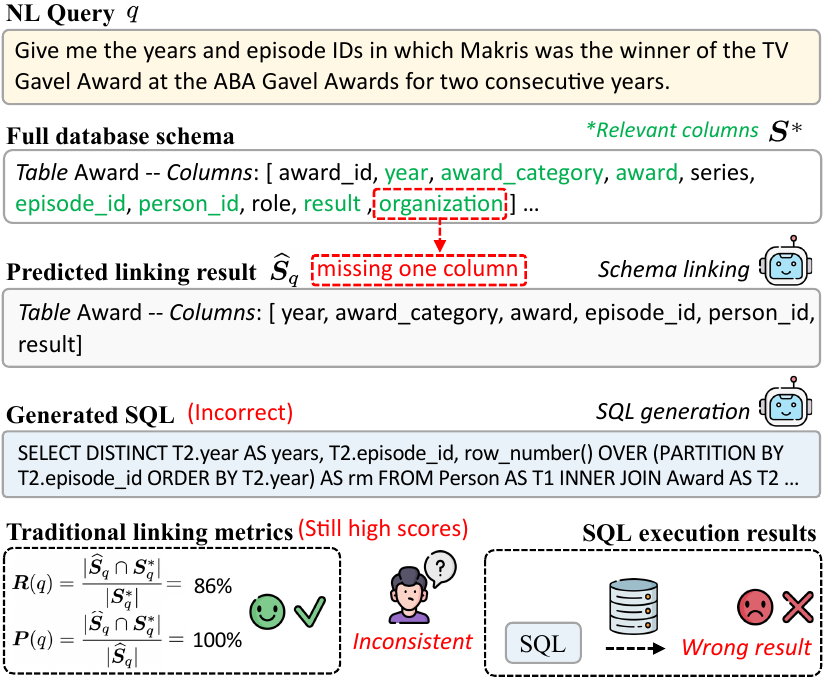}
        \caption{Commonly used metrics, like Recall and Precision, fail to accurately reflect actual schema linking performance when relevant elements are missing.
        } 
        \label{fig:intro-figure_schema-linking-metrics}
\end{figure}

Recent state-of-the-art text-to-SQL models primarily focus on the final SQL generation stage, often relying on basic or even neglecting schema linking strategies. Existing schema linking methods can generally be categorized into three types. First, some frameworks directly use the full database schema as input without a specialized schema linking mechanism, instead depending on the reasoning capabilities of LLMs. For example, DAIL-SQL~\cite{gao2023dailsql}, E-SQL~\cite{qu2024before}, and Distillery-SQL~\cite{maamari2024death} enhance LLMs’ ability to process full schema through effective prompting. Second, recall-based schema linking methods have been developed. For instance, ResdSQL~\cite{li2023resdsql} and CodeS~\cite{li2024codes} train encoder-based language models to rank and predict relevant schema elements. They employ decoupled linking strategies to separately handle tables and columns, enabling more effective use of semantic information. Third, generative schema linking methods typically leverage LLMs as schema linking models. DTS-SQL~\cite{pourreza2024dtssql} finetunes a local LLM to select relevant tables given user queries, while Solid-SQL~\cite{liu-etal-2025-solid} proposes data augmentation to improve schema linking performance. Other approaches, such as TA-SQL~\cite{qu2024before}, use in-context learning to guide LLMs in generating dummy SQL that abstracts linked schema elements. Multi-agent models like CHESS~\cite{talaei2024chess} design dedicated schema selector agents that prompt LLMs to choose relevant tables and columns. Furthermore, RSL-SQL~\cite{cao2024rsl} adopts a bidirectional approach for schema simplification, aiming to further refine schema linking.

Despite the critical role of schema linking, traditional metrics like Recall and Precision~\cite{li2023resdsql,li2024codes,qu2024before} fail to assess missing relevant elements adequately. A single missing element can render SQL generation impossible, yet current metrics may still yield high scores that misrepresent true model performance.
As shown in Fig~\ref{fig:intro-figure_schema-linking-metrics}, the schema linking result $\widehat{\mS}_q$ misses one key column, organization, which directly causes the failure to generate correct SQL due to incomplete schema linking results. However, traditional Recall and Precision metrics still give $\widehat{\mS}_q$ falsely high scores. This happens because Recall does not severely penalize the model if most relevant elements are found, allowing a high score despite missing a few relevant elements. Likewise, Precision can be unreasonably high as long as the number of correctly linked elements is large compared to the total predicted ones, regardless of how many are missed. This discrepancy skews schema linking evaluation and may hinder progress in text-to-SQL research.

To address this issue, we propose tailored schema linking metrics, \textbf{Recall$^+$ and Precision$^+$}, which incorporate a restricted missing-element indicator to more accurately assess linking performance. These metrics impose penalties when relevant elements are missing, since the absence of any relevant element in the schema linking results prevents the SQL generation model from obtaining the complete information needed to generate the correct SQL. Based on the proposed evaluation metrics, we find that existing schema linking methods face a significant issue—\textbf{the Missing \& Redundancy Seesaw Problem}: There exists a seesaw phenomenon between missing and redundant elements in current schema linking methods: Generative linking strategies tend to have fewer redundant elements but more missing ones, while recall-based models usually have fewer missing elements but more redundancy. It is difficult for current methods to achieve both low missing and low redundancy at the same time, creating a trade-off situation similar to a seesaw. Developing schema linking approaches that can effectively reduce both missing and redundant schema elements remains a significant challenge.

To address this seesaw problem, we propose the \textbf{\underline{K}n\underline{a}psack Optimization-based \underline{S}chema \underline{L}inking \underline{A}pproach (KaSLA)}, aims to link the relevant schema elements for each query while minimizing missing and redundant cases simultaneously. KaSLA introduces a binary-probabilistic score function to predict a robust relevance score for each element. In addition, KaSLA identifies redundancy of each element and determines an upper redundancy tolerance for each query by referencing its most similar queries in a pre-constructed pool of query and linking result pairs. Finally, KaSLA performs table and column linking hierarchically; it links columns based on intermediate table linking results, which reduces the candidate search space. During each linking step, KaSLA uses a 0-1 knapsack optimization approach to select the most relevant set of elements under the defined upper redundancy tolerance, aiming to avoid missing elements and minimize redundancy, all within an efficient dynamic programming framework. Extensive experiments show that KaSLA improves a text-to-SQL framework's SQL generation accuracy by simultaneously reducing missing and redundant information in schema linking.

In summary, our contributions are as follows:
\begin{itemize}[leftmargin=2em]
\item We introduce enhanced schema linking metrics, Recall$^+$ and Precision$^+$, which can accurately capture the missing of relevant elements and reflect the actual linking performance aligning with the SQL generation.
\item We propose the \underline{K}n\underline{a}psack optimization-based \underline{S}chema \underline{L}inking \underline{A}pproach (KaSLA) to link the relevant schema elements for each query while minimizing missing and redundant elements simultaneously.
\item KaSLA can serve as a general plug-in method to enhance LLM-based text-to-SQL framework by provide optimal linking results. Extensive experiments demonstrate improved performance in the SOTA text-to-SQL models when using KaSLA for schema linking.
\end{itemize}

\section{Preliminary Study}
\label{sec: Preliminaries}
In this section, we outline the definitions for the schema linking task, SQL generation task and 0-1 knapsack optimization. Additionally, we introduce the traditional schema linking metrics and then discuss their limitations.

\subsection{Definitions}
\subsubsection{Schema Linking Task}\label{sec: schema linking}
Given a natural language query $q$ and a corresponding database schema $\gS_q$ containing schema elements, tables $\gT = \{t_1, \cdots, t_{|\gT|}\}$ and columns $\gC = \{c_1, \cdots, c_{|\gC|}\}$ in each table. Schema linking is a common process used in text-to-SQL to predict the relevant schema elements (tables and columns) with $q$ from $\gS_q$:
\begin{equation}
f_{\text{linking}}: (q, \gS_q) \mapsto \widehat{\mS}_q,
\end{equation}
where $\widehat{\mS}_q$ is a subset of $\gS_q$ that includes schema element $s \in \gS_q$ that relevant to $q$ and will be used for subsequent SQL generation. Each $s$ could represent a specific table or column. Additionally, we denote $\mS_q^*$ as the ground truth linking results. 

\subsubsection{SQL Generation Task}\label{sec: text-to-SQL}
With the natural language query $q$ and the schema linking results $\widehat{\mS}_q$, the function of SQL generation task is to generate an SQL statement that is able to retrieve the pertinent information of $q$ from the database: \begin{equation}
f_{\text{sql}}: (q, \widehat{\mS}_q) \mapsto {\text{S}\widehat{\text{Q}}\text{L}}_q,
\end{equation}
where $\text{S}\widehat{\text{Q}}\text{L}_q$ represents the predicted SQL statement.

\subsubsection{0-1 Knapsack Optimization}\label{sec:0-1 Knapsack Problem}
The 0-1 knapsack problem~\cite{freville2004multidimensional} is a classic combinatorial optimization challenge. It involves selecting items from a finite item set, each with a value $\gV$ and weight $\gW$, to place into a knapsack with a maximum weight capacity $\gC$. Each item can be selected only once.
Formally, the objective of 0-1 knapsack optimization is to select items such that their total weight remains below $\gC$ while maximizing total value of these selected items:
\begin{equation} \label{equ: 0-1 Knapsack Problem}
\widehat{N} = \arg \max \sum_{i \in \widehat{N}} \gV_i, \ \text{s.t.} \sum_{i \in \widehat{N}} \gW_i \leq \gC,
\end{equation} where $\widehat{N}$ is the selected item set for the knapsack. 

\subsection{Traditional Schema Linking Metrics} \label{sec: traditional schema linking metrics}
\subsubsection{AUC} 
\begin{figure}[t]
	\centering
        \includegraphics[width=1.0\linewidth]{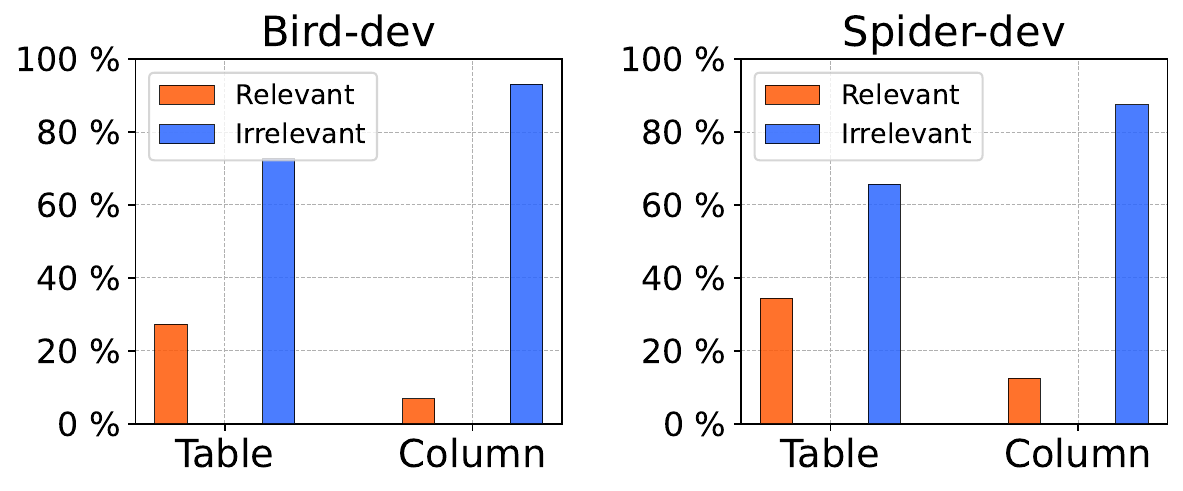}
        \caption{The imbalanced proportion of relevant schema elements with the irrelevant ones in the commonly used text-to-SQL datasets.} 
        \label{fig: Schema_Entity_M_vs_unM_Vertical_Bar}
\end{figure}

AUC is utilized to evaluate schema linking by treating it as a binary classification task, relevant elements are labeled as 1 and irrelevant ones as 0~\cite{qu2024before}. However, as depicted in Fig~\ref{fig: Schema_Entity_M_vs_unM_Vertical_Bar}, the significant imbalance between relevant and irrelevant elements will hinder the AUC from accurately reflecting the linking performance. A model can cheat its way to a high AUC by classifying many irrelevant elements, an easier task than accurately identifying the few relevant ones.

\subsubsection{Recall and Precision} Recall and Precision are commonly used in recent text-to-SQL research to evaluate schema linking performance~\cite{li2023resdsql,li2024codes,qu2024before}. For a query $q$ with the predicted schema linking result $\widehat{\mS}_q$, their formula is shown below:
\begin{equation}\label{traditional recall and precision}
\mR(q) =\frac{|\widehat{\mS}_q \cap \mS^*_q|}{|\mS^*_q|}, \mP(q) = \frac{|\widehat{\mS}_q \cap \mS^*_q|}{|\widehat{\mS}_q|},
\end{equation}
where $\mS^*_q$ is the ground truth linking result. Recall $\mR(q)$ measures the proportion of relevant elements correctly identified, specifically how many of the actual relevant elements in $\mS^*_q$ are found in $\widehat{\mS}_q$. Conversely, Precision $\mP(q)$ assesses the accuracy of predictions by determining the proportion of elements predicted as relevant in $\widehat{\mS}_q$ that are indeed correct.

As shown in Fig~\ref{fig:intro-figure_schema-linking-metrics}, the predicted schema linking result $\widehat{\mS}_q$ fails to include an essential column, ``organization". Without this information, the SQL generation model cannot generate the correct SQL statement. Nonetheless, Recall and Precision still report high scores, as they mainly reflect the number of correctly identified elements but overlook the impact of missing relevant ones. This demonstrates that these metrics can overestimate the effectiveness of schema linking, especially in this scenario where even a single omission is detrimental.

\begin{figure}[!t]
	\centering
        \includegraphics[width=1.0\linewidth]{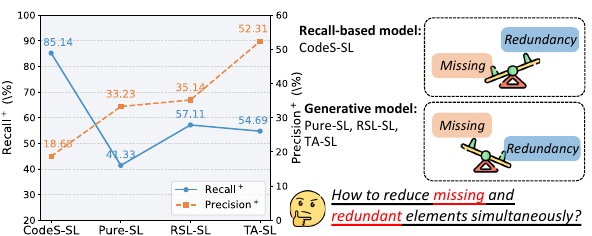}
        \caption{Missing \& redundancy seesaw problem in schema linking. We evaluate the column linking performance of schema linking baselines on BIRD-dev.} 
        \label{fig: seesaw problem}
\end{figure}

\section{Enhanced Schema Linking Metrics}
\label{sec: Enhanced Schema Linking Metrics}
In this section, we introduce enhanced schema linking metrics considering the absence of relevant elements during evaluation, thereby accurately assessing the performance of schema linking models. Addtionaly, we discuss the missing \& redundancy seesaw
problem in current schema linking methods.

\subsection{Recall$^+$ and Precision$^+$:}
To address the limitations of Recall and Precision, we firstly introduce a restricted element missing indicator to detect whether relevant elements are missing in the predicted schema linking result $\widehat{\mS}_q$, The formal definition is as follows:
\begin{equation} \label{equ: indicator function}
\bm{\mathbbm{1}}(q) = 
\begin{cases} 
1, & \text{if } \widehat{\mS}_q \cap \mS^*_q = \mS^*_q, \\
0, & \text{otherwise}.
\end{cases}\end{equation}
It takes a value of 1 when the predicted result $\widehat{\mS}_q$ contain all the elements in the ground truth $\mS^*_q$, meaning no relevant elements are missing. Conversely, it is set to 0 when there are any missing. Building on this missing indicator, we define Recall$^+$ and Precision$^+$ as follows:
\begin{equation} \label{equ: recall plus}
\mR^{\text{\textbf{+}}}(q) = \bm{\mathbbm{1}}(q)   \mR(q), \mP^{\text{\textbf{+}}}(q) = \bm{\mathbbm{1}}(q)  \mP(q).
\end{equation}
Such metrics strengthen the penalty for missing any relevant element, as schema linking results with missing are inherently incorrect for SQL generation.

\begin{figure*}[!t]
	\centering
        \includegraphics[width=1.0\linewidth]{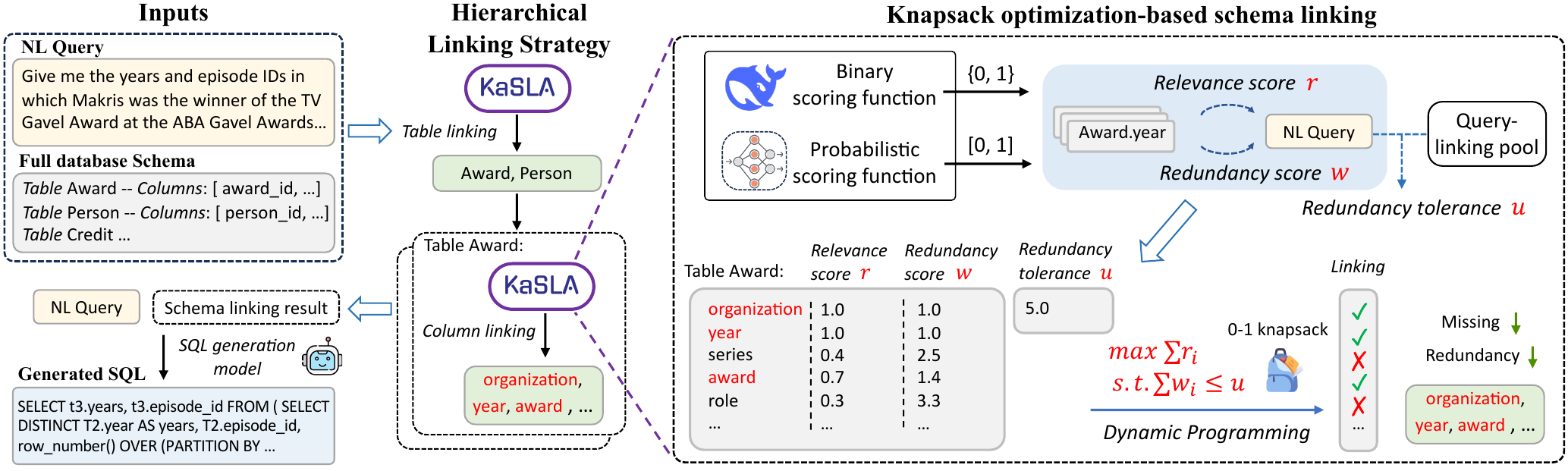}
        \caption{Overall framework of proposed knapsack optimization-based schema linking approach (KaSLA) and and how it operates within a text-to-SQL system. Traditional schema linking models often struggle with both missing relevant elements and the inclusion of redundant ones. KaSLA addresses this challenge by maximizing the total estimated relevance while maintaining total redundancy within a specified tolerance using knapsack optimization. Serving as a general plug-in method, KaSLA enhances the SQL generation of text-to-SQL models by replacing their existing schema linking processes.} 
        \label{fig: KaSLA main figure}
\end{figure*}

Since Recall$^+$ evaluates whether there are missing elements in the linking results and Precision$^+$ measures the proportion of redundant elements, we use their harmonic mean, called F1$^+$, as the overall metric. The formula is as follows:
\begin{equation} \label{equ: traditional F1 for one question}
{\mF1}^{\text{\textbf{+}}}(q) =\frac{2 \mR^{\text{\textbf{+}}}(q) \mP^{\text{\textbf{+}}}(q)} {\mR^{\text{\textbf{+}}}(q) + \mP^{\text{\textbf{+}}}(q)}.
\end{equation}
F1$^+$ combines the strengths of Recall$^+$ and Precision$^+$ and provides a comprehensive evaluation of schema linking.

Taking the same example in Fig~\ref{fig:intro-figure_schema-linking-metrics}, our enhanced metrics Recall$^+$ and Precision$^+$ offer a more reliable evaluation. When the predicted result lacks any key element, such as the ``organization" column, these metrics assign a score of zero, signaling a definite failure in schema linking for SQL generation. By strictly penalizing missing crucial information, Recall$^+$ and Precision$^+$ ensure models are only rewarded for comprehensive and complete predictions, thus aligning the metric outcomes closely with actual SQL generation success.

\subsection{The Missing \& Redundancy Seesaw Problem}
After eliminating the misleading effects caused by missing elements, Recall$^+$ and Precision$^+$ can be used to accurately measure a schema linking model's ability to link relevant elements and avoid redundant elements, respectively. We applied several baselines on BIRD-dev and evaluated their column linking preferences. As shown in Fig~\ref{fig: seesaw problem}, we observed a significant seesaw phenomenon between missing and redundancy in these schema linking methods.

For example, recall-based models like CodeS-SL rarely miss relevant columns, resulting in high Recall$^+$, but it includes too many redundant columns, leading to very low Precision$^+$. In contrast, generative methods such as Pure-SL, RSL-SL and TA-SL tend to have more missing columns, resulting in lower Recall$^+$, but their linking results contain fewer redundant columns, yielding relatively higher Precision$^+$ compared to the recall-based model, CodeS-SL.

This prominent missing \& redundancy seesaw problem inspires a key question in schema linking: How can we minimize both element missing and element redundancy while ensuring that relevant elements are properly linked? To address this issue, we propose the Knapsack Optimization-based Schema Linking Approach (KaSLA), which uses a knapsack optimization framework to improve the schema linking process. The details are provided in Section~\ref{sec:Knapsack Optimization-based Schema Linking Approach (KaSLA)}.

\section{Knapsack Optimization-based Schema Linking Approach (KaSLA)} \label{sec:Knapsack Optimization-based Schema Linking Approach (KaSLA)}

In this section, we present KaSLA, a novel schema linking framework for text-to-SQL tasks that explicitly addresses both missing and redundant element challenges through a knapsack optimization approach. We first provide an overview of the KaSLA framework in Section~\ref{sec: Overall framework}, outlining its key components, optimization objective and hierarchical structure. Next, we introduce the core components in Section~\ref{sec: Key components}, including the hybrid relevance estimation mechanism, redundancy assessment strategy, and the formulation of redundancy tolerance for each query. We then describe how schema linking works as a knapsack optimization process and explain the hierarchical linking strategy for efficient and precise schema element selection in Section~\ref{sec: Schema Linking with Knapsack Optimization}. Each subsection details a critical aspect of KaSLA, jointly enabling accurate and efficient schema linking for robust text-to-SQL generation.

\subsection{Overall Framework}\label{sec: Overall framework}
As illustrated in Fig~\ref{fig: KaSLA main figure}, KaSLA is designed to systematically address the challenges of both missing and redundant schema elements in text-to-SQL systems by using a knapsack optimization approach. The overall process begins with two key inputs: a natural language query and the full database schema. As visible in the left portion of Fig~\ref{fig: KaSLA main figure}, KaSLA follows a hierarchical linking strategy, where it first performs table linking to identify tables relevant to the query, and then proceeds to link columns within each selected table. This approach effectively reduces the candidate space and enhances efficiency when working with complex database schema.

At the core of KaSLA's schema linking is a hybrid scoring module that combines a binary scoring function (yielding hard 0/1 relevance decisions) and a probabilistic scoring function (providing soft relevance probabilities) for each schema element. As shown in the right side of Fig~\ref{fig: KaSLA main figure}, these scores are integrated to assign a final relevance score~$r$ to every candidate element. In parallel, each element receives a redundancy score~$w$, and for each query, a redundancy tolerance~$u$ is estimated by analyzing similar queries with their linking results from a pre-constructed query-linking pool, as depicted in the figure. This step ensures the model adapts its tolerance to different query types and schema complexities.

The selection of relevant schema elements is then formulated as a 0-1 knapsack problem: KaSLA seeks to maximize the total relevance scores of linked elements while keeping the sum of their redundancy scores within the estimated tolerance $u$. The optimization objective, shown in red in Fig~\ref{fig: KaSLA main figure}, is solved efficiently using a dynamic programming algorithm. The knapsack selection process ensures that the final set of linked schema elements contains all necessary information for SQL generation while minimizing unnecessary redundancy.

After schema linking, the selected elements are passed to the downstream SQL generation model, which constructs the final SQL statement in accordance with the linking result. As shown on bottom left of Fig~\ref{fig: KaSLA main figure}, this process allows KaSLA to serve as a plug-in method for existing text-to-SQL frameworks, helping them avoid common pitfalls of missing relevant elements or introducing redundant ones. By leveraging hierarchical optimization, hybrid scoring and redundancy-aware selection, KaSLA robustly improves schema linking across diverse and large-scale text-to-SQL applications.

\subsection{Key Factor Estimation}\label{sec: Key components}
In this section, we present a detailed introduction to the estimation functions of key factor in KaSLA: element relevance, element redundancy and upper redundancy tolerance.
\subsubsection{Relevance Estimation}\label{sec:Relevance Estimation}
To precisely assess the relevance of each schema element $s$ in relation to the given query $q$, we employ a hybrid binary-probabilistic score function. This function integrates a generative model for binary scoring alongside an encoding model for probabilistic scoring, allowing for a comprehensive estimation of relevance.

\paragraph{Binary Scoring Function}
This component is tasked with providing binary relevance assessments, determining whether a schema element is relevant or not in a straightforward manner with hard 0/1 relevance decisions:
\begin{equation}\label{eq. Binary Scoring Function}
f_{\text{binary}}: (q, \gS_q) \mapsto \{r^b_1, \cdots, r^b_{|\gS_q|}\},
\end{equation}
where $r^b_i \in \{0, 1\}$ represents the binary score assigned to the $i$-th element $s_i \in \gS_q$. A score of 1 indicates relevant while 0 signifies redundant. 

Considering efficiency, we finetune a lightweight LLM, DeepSeek-Coder-1.3B~\cite{guo2024deepseek}, with a well-established parameter-efficient technique, LoRA~\cite{hu2021lora}. We employ a strategy whereby the initial generation of a simulated SQL provides a contextual foundation for the subsequent generation of schema linking. The loss function during fine-tuning is defined as follows:
\begin{equation}
\mathcal{L}_{\text{binary}} = - \sum_{q \in Q} \log P({\text{SQL}}^*_q, \mS_q^* \mid q, \gS_q),
\end{equation} where $Q$ is the set of queries in the training dataset. ${\text{SQL}}^*_q$ and $\mS^*$ are the ground truth SQL and linking result, respectively. This formulation utilizes a joint generation of a simulated SQL and the schema linking result, thereby ensuring that the linking generation benefits from the contextual alignment provided by the simulated SQL during both fine-tuning and inference. Noted that the simulated SQL generated during this process will not be utilized in subsequent steps.

\paragraph{Probabilistic Scoring Function}
In conjunction with the binary scoring model, we introduce a probabilistic scoring model to enhance prediction robustness. The primary motivation is to mitigate the risk of the binary model misclassifing relevant elements. The probabilistic scoring model assigns a soft score to each element as the linking probability to reduce the likelihood of the missing of potential relevant elements, thereby supporting the binary model. The equation is defined as follows:
\begin{equation}\label{eq. Probabilistic Scoring Function}
f_{\text{prob.}}: (q, \gS_q) \mapsto \{r^p_1, \cdots, r^p_{|\gS_q|}\},
\end{equation}
where  $r^p_i \in [0, 1]$ represents the probabilistic score assigned to the $i$-th element $s_i \in \gS_q$. A value closer to 1 indicates that the element is considered more relevant to query q.

For the probabilistic scoring model, we utilize an encoding model, RoBERTa-Large~\cite{liu2019roberta}, to derive the semantic embeddings for the query and all schema elements. Following the methodology of ~\cite{li2023resdsql,li2024codes}, we employ a cross-attention network to jointly embed the semantic representations of tables and their columns. The dot product of the query embedding and the element embedding yields the final probabilistic score. This probabilistic model is trained using the presence of $s$ in $\mS^*$ as the learning objective. The loss function is defined as follows:
\begin{equation}\label{eq:training loss of recall model}
\mathcal{L}_{\text{prob.}} = \sum_{q \in Q} \text{FL}(\mathbbm{1}(s \in \mS^*_q), f_{\text{prob.}}(q,\gS_q)),
\end{equation}
where $\text{FL}(\cdot)$ denotes the focal loss function ~\cite{ross2017focal}, designed to emphasize learning from hard negative samples.


\paragraph{Relevance Scoring Estimation}
The final function for estimating the relevance of each schema element integrates the scores derived from both the binary and probabilistic functions:
\begin{equation}\label{eq. Relevance scoring estimation}
\begin{split}
    f_{\text{relevance}}:& (q, \gS_q) \mapsto \{r_1, \ldots, r_{|\gS_q|}\} \\
    & = \left\{\min(1, r^b_i + r^p_i) \mid s_i \in \gS_q \right\},
\end{split}
\end{equation}
where $ r_i$ denotes the relevance score assigned to the $i$-th schema element $s_i \in \gS_q$ given the query $q$. An upper limit of 1 is imposed on $r_i$ to prevent excessively strong contrasts between potentially relevant elements, as each relevant schema element is equally important for accurate SQL generation. We use ${\gR}_{\gS_q} = \{r_1, \ldots, r_{|\gS_q|}\}$ to denote the relevance score set of $\gS_q$.

\subsubsection{Redundancy Estimation}
To minimize the inclusion of redundant elements during schema linking, we define both a redundancy score for each schema element $s \in \gS_q$ and the upper redundancy tolerance for query $q$.

\paragraph{Redundancy Scoring Estimation}
Recognizing that elements with lower relevance scores are more likely to be redundant, whereas those with higher scores are more essential, we assign a redundancy score $w_i$ to each schema element $s_i \in \gS_q$ in relation to query $q$. This score is calculated as the inverse of its relevance score $r_i$, formulated as follows:
\begin{equation}\label{eq. Redundancy Scoring Estimation}
\begin{split}
    f_{\text{redundancy}}:& (q, {\gR}_{\gS_q}) \mapsto \{w_1, \ldots, w_{|\gS_q|}\} \\
    & = \left\{{r_i}^{-1} \mid r_i \in {\gR}_{\gS_q} \right\}.
\end{split}
\end{equation}
It ensures elements with lower relevance scores contribute more to redundancy, thereby effectively prioritizing more relevant schema elements. We use ${\gW}_{\gS_q} = \{w_1, \ldots, w_{|\gS_q|}\}$ to denote the redundancy score set of $\gS_q$. We provide a detailed discussion of the designed inverse relation between redundancy and relevance scores in $f_{\text{redundancy}}$ in Section~\ref{ablation_Redundancy Estimation}.

\paragraph{Redundancy Tolerance}
Upper tolerance of redundancy defines the limit on the total redundancy score of elements in the schema linking results, as an excess of redundant elements can significantly disrupt subsequent SQL generation. Estimating a suitable upper redundancy tolerance during inference is challenging. For the SQL generation of some queries, a lenient tolerance may be beneficial since it allowing the inclusion of more potentially relevant elements in the linking results to prevent missing. However, for other queries, a lenient tolerance could introduce too many redundant elements, which might significantly disrupt the final SQL generation process.

To address this challenge, we hypothesize that similar queries have comparable tolerance levels for redundancy. We sample queries from the training dataset where ground truth linking results are available, and use these paired samples to construct a query-linking pool. During inference, when predicting the redundancy tolerance for a given query \(q\), we first retrieve the top-\(K\) most similar queries from the query-linking pool using sentence similarity measures~\cite{gao2021simcse}. For a query \(q_k\) in the retrieved top-\(K\) queries, the redundancy score set for its ground truth linking results $\gS^{*}_{q_k}$ can be denoted as ${\gW}_{\gS^{*}_{q_k}}$. We sum the total ${\gW}_{\gS^{*}_{q_k}}$ and use it as the upper redundancy tolerance of $q_k$. The maximum of these upper redundancy tolerances of the \(K\) most similar queries from the query-linking pool is then used as the redundancy tolerance estimate for query \(q\):
\begin{equation}\label{eq. Upper Redundancy Tolerance}
f_{\text{tolerance}}: (q) \mapsto  u_{q} = \max_{q_k \in \text{TopK}(q)}  \sum_{w_j \in {\gW}_{\gS^{*}_{q_k}}} w_j.
\end{equation}
By capturing the maximum redundancy tolerances of the most similar queries, this approach effectively defines an upper bound on redundancy tolerance that is both efficient and adaptable, ensuring that while redundant elements are excluded, the risk of missing relevant elements is minimized. The ablation studies on $f_{\text{tolerance}}$ are detailed in Section\ref{ablation_Redundancy Tolerance Estimation}.

\subsection{Schema Linking with Knapsack Optimization}\label{sec: Schema Linking with Knapsack Optimization}
Given (i) the relevance score $r_i$ and redundancy score $w_i$ for each candidate element $s_i \in \gS_q$, and (ii) the upper redundancy tolerance $u_q$ defined at the query level, we propose a hierarchical schema linking strategy incorporating knapsack optimization. This approach is designed to ensure the linking of relevant schema elements to avoid missing and minimize the inclusion of redundant ones for a precise schema linking.

\subsubsection{Knapsack Optimization}
The 0-1 knapsack problem~\cite{freville2004multidimensional} is a classic combinatorial optimization problem. It involves selecting items from a finite set. Given a set of items, each with a \textit{value} and a \textit{weight}, the goal is to select a subset of items that maximizes the total value while ensuring the total weight does not exceed a limited knapsack \textit{capacity}. In the ``0-1'' variant, each item is either fully included or excluded (i.e., no fractional selections are permitted).

Inspired by this, we develop a knapsack optimization-based schema linking strategy to ensure the inclusion of relevant schema elements while minimizing redundant ones, thus ach ieving optimal schema linking. Formally, the optimization objective are defined as follows:
\begin{equation}
\begin{split}
    f_{\text{Knap}}:& (q, \gS_q) \mapsto \widehat{\mS}_q \\&  = \arg\max \sum_{s_i \in \widehat{\mS}_q} r_i ,\quad  \text{s.t.}  \sum_{s_i \in \widehat{\mS}_q} w_i \leq u_q .
\end{split}
\end{equation}
This formulation aims to maximize the total relevance scores of the linked schema elements with a total redundancy constraint. The constraint ensures that the cumulative redundancy score remains within the upper redundancy tolerance of query $q$. To tackle this optimization problem and achieve optimal linking results, we employ an efficient dynamic programming approach. The details of this algorithm, as implemented in KaSLA, are presented in Algorithm~\ref{alg: Dynamic Programming-based 0-1 Knapsack Optimization Algorithm}. We discuss the complexity analysis of the whole knapsack optimization framework and the inference time cost of each key component in Section~\ref{Complexity Analysis and Inference Time Cost}.

\begin{algorithm}[!h]
\caption{Dynamic Programming-based 0-1 Knapsack Optimization Algorithm}
\label{alg: Dynamic Programming-based 0-1 Knapsack Optimization Algorithm}
    \KwIn{relevance score set $\gR$,  redundancy score set $\gW$, the upper redundancy tolerance $u$}
    \KwOut{schema linking set $\hat{S}$}
        $n \gets |\gR|$, $\hat{S} \gets \{\}$ \\
        $\mathcal{A} \gets \text{array of }(n+1) \times (u+1), \text{ initialized to } 0$ \\
        $ keep \gets \text{array of }(n+1) \times (u+1), \text{ initialized to False} $\\
        \For{$i \gets 1$ \textbf{to} $n$}{
            \For{$w \gets 0$ \textbf{to} $u$}{
                \eIf{$\gW[i-1] \leq w$}{
                    \eIf{$(\gR[i-1] + \mathcal{A}[i-1][w-\gW[i-1]]) > \mathcal{A}[i-1][w]$}{
                        $\mathcal{A}[i][w] \gets \gR[i-1] + \mathcal{A}[i-1][w-\gW[i-1]]$ \\
                        $keep[i][w] \gets \text{True}$ 
                    }{
                        $\mathcal{A}[i][w] \gets \mathcal{A}[i-1][w]$ 
                    }
                }{
                    $\mathcal{A}[i][w] \gets \mathcal{A}[i-1][w]$
                }
            }
        }
        $k \gets u$ \
        \For{$i \gets n$ \textbf{downto} $1$}{
            \If{$keep[i][k]$}{
                    $\hat{S}.\text{add}(i-1)$ \\
                    $k \gets k - \gW[i-1]$
                }
        }
        \Return{$\hat{S}$}
\end{algorithm}

\subsubsection{Hierarchical Linking Strategy}
To ensure computational efficiency, KaSLA adopts a hierarchical schema linking strategy consisting of two stages: (1) table linking followed by (2) column linking within selected tables. This hierarchical approach significantly reduces the search space of candidate columns, enabling KaSLA to efficiently handle large-scale databases while maintaining robust performance.

In the initial phase for a given query $q$ and the corresponding $\gS_q$, KaSLA aims to identify the relevant tables from the set of all available tables in schema $\gS_q$, denoted as $\gT_q$:
\begin{equation}
\widehat{\mT}_q  = f_{\text{Knap}}({\gR}_{\gT_q}, {\gW}_{\gT_q}, u^{\gT}_q ),
\end{equation}
where ${\gR}_{\gT_q}$ and ${\gW}_{\gT_q}$ are the sets of estimated relevance and redundancy scores for tables in schema $\gS_q$, and $u^{\gT}_q$ is the upper redundancy tolerance for table linking. Different from the general schema elements $\gS_q$ used in the equations in Section~\ref{sec: Key components}, the notation $\gT_q$ and $\gT$ specifically refer to tables in schema linking. After determining the selected tables, for each $t \in \widehat{\mT}_q$, let $\gC^t_q$ represent the columns in table $t$. KaSLA then optimizes column linking within each selected table:
\begin{equation}
\widehat{\mC}^t_q  = f_{\text{Knap}}({\gR}_{\gC^t_q}, {\gW}_{\gC^t_q}, u^{\gC}_q),
\end{equation}
where ${\gR}_{\gC^t_q}$ and ${\gW}_{\gC^t_q}$ are the sets of estimated relevance and redundancy scores for columns in table $t$, and $u^{\gC}_q$ is the upper redundancy tolerance for column linking. The final schema linking result $\widehat{\mS}_q$ is the union of all selected tables and their corresponding selected columns, represented as:
\begin{equation}
\widehat{\mS}_q = \widehat{\mT}_q \cup \bigcup_{t \in \widehat{\mT}_q} \widehat{\mC}^t_q.
\end{equation}
We present the complete procedure of the hierarchical linking strategy in Algorithm~\ref{alg:Hierarchical KaSLA Strategy}.

\begin{algorithm}[thbp]
	\caption{Hierarchical Linking Strategy}
	\label{alg:Hierarchical KaSLA Strategy}
	\KwIn{user query $q$, full schema $\gS_q$}
	\KwOut{schema linking results $\widehat{\mS}_q$, including a table linking set $\widehat{\mT}_q$ and the corresponding column linking set $ \widehat{\mC}^t_q$ for each table $t \in \widehat{\mT}_q$}
        $ ({\gR}_{\gT_q}, \left\{{\gR}_{\gC^t_q}  \right\}_{t \in \gT_q} )= f_{\text{relevance}}(q, \gS_q)$, \\
        $ {\gW}_{\gT_q} = f_{\text{redundancy}}(q, {\gR}_{\gT_q}), (u^{\gT}_q, u^{\gC}_q )= f_{\text{tolerance}}(q) $ \\
        $ \widehat{\mT}_q  = f_{\text{Knap}}({\gR}_{\gT_q}, {\gW}_{\gT_q}, u^{\gT}_q )$ \tcp*[f]{\scriptsize\spaceskip=0.3em Table linking with Alg.~\ref{alg: Dynamic Programming-based 0-1 Knapsack Optimization Algorithm}}\\
	\For{$t \in \widehat{\mT}_q $}{
             $ {\gW}_{\gC^t_q} = f_{\text{redundancy}}(q, {\gR}_{\gC^t_q})$ \\
             $ \widehat{\mC}^t_q  = f_{\text{Knap}}({\gR}_{\gC^t_q}, {\gW}_{\gC^t_q}, u^{\gC}_q)$\tcp*[f]{\scriptsize\spaceskip=0.3em Column linking with Alg.~\ref{alg: Dynamic Programming-based 0-1 Knapsack Optimization Algorithm}} \\
	}
        \Return{$\widehat{\mS}_q = \widehat{\mT}_q \cup \bigcup_{t \in \widehat{\mT}_q} \widehat{\mC}^t_q$}
\end{algorithm}

\section{Experiments}
\label{sec:Experiments}

We conducted comprehensive experiments to evaluate KaSLA and address the following research questions: \textbf{RQ1:} Does KaSLA enhance existing text-to-SQL baselines by improving their schema linking capabilities? \textbf{RQ2:} Does KaSLA demonstrate superior schema linking performance by handling element missing and redundancy? \textbf{RQ3:} Does each component in KaSLA contribute to the overall performance, and what is the effect of parameter scaling? \textbf{RQ4:} Is KaSLA an efficient model for real-world text-to-SQL applications, especially in terms of its ability to transfer to unseen domains?

\subsection{Experiment Setup}\label{sec: Experiment Setup}

\subsubsection{Datasets and Evaluation Metrics}
We conducted experiments on two well-known large-scale text-to-SQL datasets, BIRD~\citep{li2023BIRD} and Spider~\citep{yu2018Spider}. Both datasets feature human-annotated queries and SQLs, complex database elements, and challenging cross-domain scenarios meticulously.
For text-to-SQL evaluation, we measured performance using two well-established execution-based metrics: Execution Accuracy (EX)~\citep{yu2018Spider,li2023BIRD} and Valid Efficiency Score (VES)~\citep{li2023BIRD}.

\subsubsection{Baselines}
We incorporate a comprehensive set of competitive baselines covering both LLMs with in-context learning and with fine-tuning. For the in-context learning category, we include C3-SQL~\cite{dong2023c3}, DIN-SQL~\cite{pourreza2023dinsql}, MAC-SQL~\cite{wang2024macsql}, DAIL-SQL~\cite{gao2023dailsql}, SuperSQL~\cite{li2024dawn}, Dubo-SQL~\cite{thorpe2024dubo}, TA-SQL~\cite{qu2024before}, E-SQL~\cite{caferouglu2024sql}, CHESS~\cite{talaei2024chess}, and RSL-SQL~\cite{cao2024rsl}. For the fine-tuning category, we include DTS-SQL~\cite{pourreza2024dtssql}, TA-SQL~\cite{qu2024before}, and CodeS~\cite{li2024codes}. We also introduce Pure-SL, which fine-tunes LLMs for schema linking without specialized mechanisms. For baseline methods with a modular schema linking component, we denote variants referring specifically to the schema linking module with an '-SL' suffix (e.g., CodeS-SL).

\subsubsection{Implementation Details}\label{sec: Implementation Details}
We evaluate KaSLA and all baselines under comparable settings. For most in-context learning baselines, we use GPT-4-turbo. For E-SQL, CHESS, and RSL-SQL, we use deepseek-V3 instead of their original GPT-4o due to API cost, as deepseek-V3 offers similar performance~\cite{liu2024deepseek}. For open-source LLM experiments, CodeS-15B~\citep{li2024codes} and StarCoder2-15B~\citep{lozhkov2024starcoder} serve as the base models, and all baselines are reproduced using LoRA~\citep{hu2021lora} for parameter-efficient fine-tuning with learning rate $1e-4$, cosine decay, batch size 16, and 3 training epochs. For KaSLA, the top-$K$ for redundancy tolerance is set to 30, and we fine-tune RoBERTa-Large~\citep{liu2019roberta} as the probabilistic scoring model and Deepseek-coder-1.3B~\cite{guo2024deepseek} as the binary scoring model, both using the same fine-tuning setup as the baselines.

\begin{table*}[t]
   \caption{The SQL generation performance of enhanced text-to-SQL models using KaSLA, which features 1.6 B parameters combining DeepSeek-coder-1.3B and RoBERTa-Large, is evaluated in terms of Execution Accuracy (EX) (\%) and Valid Efficiency Score (VES) (\%) on the BIRD-dev and Spider-dev datasets. We use KaSLA to replace the original schema linking component or the full schema in the baseline models, denoted as ``+ KASLA'', and the \colorbox{gray!20}{relative improvements} are highlighted with a gray background.}
   \label{tb:main_results}
   \begin{center}
   \resizebox{\linewidth}{!}{
   \scalebox{1}{
   \begin{tabular}{c|c|ccc|c|c|cccc|c|cc}
   \toprule[1pt]
   \multirow{4}{*}{\makecell[c]{Base Model}}& \multirow{4}{*}{\makecell[c]{Method}} &\multicolumn{5}{c|}{BIRD-dev} &\multicolumn{6}{c}{Spider-dev} \\
   \cmidrule(lr){3-7}\cmidrule(lr){8-13}
   & &\multicolumn{4}{c|}{EX} &  VES     &\multicolumn{5}{c|}{EX} &  VES \\
   \cmidrule(lr){3-7}\cmidrule(lr){8-13}
   & &Easy&Medium & Hard & Total &   Total                              &Easy &Medium &Hard &Extra &Total&Total \\
   \midrule
   \multirow{7}{*}{\makecell[c]{GPT-4}}& MAC-SQL   &-  &-  &-  &57.56  &  58.76                             &- &-  &-  &-    &86.75    &-          \\
    &DIN-SQL   &-  &-  &-  &50.72    & 58.79                             &92.34  &87.44  &76.44  &62.65  &82.79  &81.70         \\
    &DAIL-SQL   &62.49  &43.44  &38.19  &54.43      & 55.74                   &91.53  &89.24  &77.01  &60.24  &83.08   &83.11          \\
    &DAIL-SQL (SC)  &63.03  &45.81  &43.06  &55.93     & 57.20                &91.53  &90.13  &75.29  &62.65  &83.56  &-          \\
    &TA-SQL   &63.14 &{48.82} &36.81 &56.32               &  -               &93.50 &90.80 &77.60 &64.50 &85.00         & -        \\
    &SuperSQL  &{66.92}  &46.67  &{43.75}  &58.60   & 60.62       &{94.35} &{91.26}  &{{83.33}}  &{{68.67}}  &{87.04} &{85.92}  \\
   & Dubo-SQL   &-  &-  &-  &{59.71} & {66.01}            &- &-  &-  &-  &-  & -           \\
   \midrule
     \multirow{2}{*}{\makecell[c]{CodeS-15B}}  & CodeS  &65.19&48.60&38.19&57.63   & 63.22 &94.76&91.26  &72.99  &{64.46}   &84.72   & 83.52     \\
     & \textbf{CodeS   + KaSLA}      &68.32&50.75&38.19&60.17 \colorbox{gray!20}{\scriptsize $\uparrow$ 4.41\%}  &64.52 \colorbox{gray!20}{\scriptsize $\uparrow$ 2.06\%}    &96.37&89.46&76.44&63.86&84.82 \colorbox{gray!20}{\scriptsize $\uparrow$ 0.12\%}    & 84.46 \colorbox{gray!20}{\scriptsize $\uparrow$ 1.13\%}     \\
     \midrule
   \multirow{5}{*}{\makecell[c]{StarCoder2\\-15B}}   &TA-SQL  &  59.89&45.38&34.72&53.13   & 60.51            &95.56&92.38&78.16&63.25& 85.98 &  85.03       \\ 
     & \textbf{TA-SQL  + KaSLA}      &66.59  & 49.46  & 38.89 & 58.80 \colorbox{gray!20}{\scriptsize $\uparrow$ 10.67\%}   &63.56 \colorbox{gray!20}{\scriptsize $\uparrow$ 5.04\%}    &94.76&92.15&77.59&66.87&86.27 \colorbox{gray!20}{\scriptsize $\uparrow$ 0.34\%} &85.47 \colorbox{gray!20}{\scriptsize $\uparrow$ 0.52\%}  \\
    &  DTS-SQL &  63.03&46.02&34.72&55.22 &64.17 & 91.94&90.58&78.74&66.87&85.11  &  85.49  \\ 
    &  \textbf{ DTS-SQL  + KaSLA}      &63.89 &50.11   &39.58   &57.43 \colorbox{gray!20}{\scriptsize $\uparrow$ 4.00\%}  &65.20 \colorbox{gray!20}{\scriptsize $\uparrow$ 1.61\%}     &94.76&90.81&\textbf{80.46}&\textbf{68.07}&86.36 \colorbox{gray!20}{\scriptsize $\uparrow$ 1.47\%}&85.81 \colorbox{gray!20}{\scriptsize $\uparrow$ 0.37\%} \\
     \midrule
    \multirow{7}{*}{\makecell[c]{Deepseek\\-V3}} & E-SQL  &67.24  &51.61  &41.67 &60.10   & 62.16   &95.57&91.93&72.99&64.46& 85.21 &    83.81   \\ 
   & \textbf{ E-SQL + KaSLA}  &69.40   &53.55 &43.75  &62.18 \colorbox{gray!20}{\scriptsize $\uparrow$ 3.46\%}  &65.32 \colorbox{gray!20}{\scriptsize $\uparrow$ 5.08\%}   &95.97 &92.38 & 73.56 & 65.06 &85.69 \colorbox{gray!20}{\scriptsize $\uparrow$ 0.56\%}& 85.72 \colorbox{gray!20}{\scriptsize $\uparrow$ 2.28\%}   \\
    &  CHESS   &67.89  &52.90 &46.53 & 61.34  &65.28 & 95.57&92.16&73.56& 65.06& 85.50  & 85.33   \\ 
    & \textbf{  CHESS + KaSLA}   &69.94   &54.84 & 47.92 &63.30 \colorbox{gray!20}{\scriptsize $\uparrow$ 3.20\%}   &67.21 \colorbox{gray!20}{\scriptsize $\uparrow$ 2.96\%}  &96.37&92.83&74.14&65.66 &86.17 \colorbox{gray!20}{\scriptsize $\uparrow$ 0.78\%} & 86.35 \colorbox{gray!20}{\scriptsize $\uparrow$ 1.20\%} \\
    &  RSL-SQL   &69.95  &54.62 &47.22 & 63.17  &67.11 & 96.37&93.05&74.71& 66.87& 86.56  & 86.42   \\ 
    & \textbf{  RSL-SQL + KaSLA}   &\textbf{71.03}   &\textbf{56.34} & \textbf{51.39} &\textbf{64.73} \colorbox{gray!20}{\scriptsize $\uparrow$  2.47\%}   &\textbf{68.12} \colorbox{gray!20}{\scriptsize $\uparrow$ 1.35\%}  &\textbf{97.18}&\textbf{93.50}&75.29&\textbf{68.07} &\textbf{87.14} \colorbox{gray!20}{\scriptsize $\uparrow$ 1.50\%} & \textbf{87.36} \colorbox{gray!20}{\scriptsize $\uparrow$ 1.09\%} \\
   \bottomrule[1pt]
   \end{tabular}}}
   \end{center}
   \end{table*}

\subsection{Main Results of SQL Generation (\textbf{RQ1})}

To address \textbf{RQ1}, we conducted experiments to evaluate the SQL generation capabilities of state-of-the-art (SOTA) baselines enhanced by our KaSLA, with results presented in Table~\ref{tb:main_results}. KaSLA serves as a plug-in method: for models with built-in schema linking modules, it replaces their original linking component; for models without a schema linking component, it substitutes the conventional full schema prompting approach.

\begin{table*}[t]
   \caption{Schema linking performance with Recall$^+$, Precision$^+$, and F1$^+$ on BIRD-dev. We extracted the schema linking components from the baselines and used ``-SL'' instead of the original ``-SQL'' to indicate this.}
   \label{tb: Schema linking performance on BIRD-dev.}
   \begin{center}
   \resizebox{\linewidth}{!}{
   \scalebox{1}{
   \begin{tabular}{c|cc|c|c|c|c|c|c|c|c|c|c} 
   \toprule[1pt]
   \multirow{2}{*}{Dataset}&\multicolumn{2}{c|}{\multirow{2}{*}{Metric}} &\makecell[c]{Recall-based Method} &\multicolumn{8}{c|}{Generative Method} &\multirow{2}{*}{KaSLA} \\
   \cmidrule{4-12}
   &&&CodeS-SL& ICL &DIN-SL & DTS-SL & CHESS-SL & E-SL& Pure-SL& TA-SL& RSL-SL&  \\
   \midrule
  \multirow{7}{*}{BIRD-dev} 
    &\multirow{2}{*}{Recall$^+$} 
    & Table &\textbf{99.15} &88.40  &82.66 &64.99 &80.70 &69.04 &64.99 &79.47     &92.57 &94.52 \\
    &&Column &\textbf{85.14} &22.03 &16.17  &62.52   &14.93 &22.36 &41.33 &54.69   &57.11 &81.29   \\
    \cmidrule{2-13}
    &\multirow{2}{*}{Precision$^+$} 
    & Table  &43.03 &82.75 &78.80 &60.29  &49.92 &68.01  &60.29 &74.49   &\textbf{83.89} &75.00    \\
    &&Column  &18.65 &20.69&15.12 &16.50   &7.09 &21.57  &37.23 &52.31  &35.14 &\textbf{53.53}   \\
    \cmidrule{2-13}
    &\multirow{2}{*}{F1$^+$} 
    & Table  &58.21 &84.79  &80.18 &61.94   &58.51 &68.39 &61.94 &76.16   &\textbf{86.93} &81.92  \\
    &&Column  &29.56 &21.20  &15.51  &23.71   &9.03 &21.87 &38.77 &53.20   &40.60 &\textbf{62.78}    \\
   \bottomrule[1pt]
   \end{tabular}}}
   \end{center}
   \end{table*}

The results in Table~\ref{tb:main_results} demonstrate that incorporating KaSLA consistently and significantly improves core metrics, including Execution Accuracy (EX) and Valid Efficiency Score (VES), across diverse baselines and backbone models such as GPT-4, CodeS-15B, StarCoder2-15B, and Deepseek-v3. For example, on the challenging BIRD-dev dataset, KaSLA raises the EX of RSL-SQL (Deepseek-v3) and CHESS by notable margins, and also leads to substantial gains for “CodeS + KaSLA” and “TA-SL + KaSLA” with their respective local LLMs. On Spider-dev, where the schema is less complex, KaSLA still provides consistent improvements. These results highlight KaSLA’s robustness and broad applicability, making it a valuable plug-in for enhancing text-to-SQL performance in various scenarios.

\subsection{Experimental Results of Schema Linking (\textbf{RQ2})} \label{sec: Schema Linking Results} 

To answer \textbf{RQ2}, we evaluate schema linking methods on BIRD-dev using the proposed schema linking metrics, as reported in Table~\ref{tb: Schema linking performance on BIRD-dev.}. Most models perform well on table linking, but column linking remains challenging due to the large candidate space. Recall-based approaches like CodeS-SL achieve high Recall$^+$ but suffer from low Precision$^+$, especially at the column level, resulting in lower F1$^+$ scores. Generative models such as RSL-SL offer higher precision but miss more relevant elements, and even top performing methods in end-to-end SQL generation like E-SQL and CHESS show poor standalone schema linking performance. In contrast, KaSLA achieves a strong balance: its Recall$^+$ is close to the recall-based method, while its Precision$^+$ surpasses generative baselines, leading to the best overall F1$^+$ on column linking and strong table-level results. These results demonstrate that KaSLA’s knapsack-optimization framework provides the best trade-off between missing and redundant elements, making it effective for difficult column linking in text-to-SQL tasks.

\subsection{Ablation Study (\textbf{RQ3})}
To answer RQ3, we conduct comprehensive ablation studies on KaSLA, including: (i) analysis of its key components, (ii) the influence of different language model choices for the probabilistic scoring model, (iii) the effects of scaling the binary scoring model, the design of (iv) the redundancy estimation $f_{\text{redundancy}}$, and (v) the redundancy tolerance estimation $f_{\text{tolerance}}$. Unless otherwise specified, all experiments are performed with CodeS-15B as the text-to-SQL backbone.
   
\begin{table}[!t]
   \caption{Ablation studies of the key components of KaSLA. We conduct the experiments for CodeS + KaSLA on BIRD-dev and use CodeS-15B as the text-to-SQL backbone.}
\label{tb: ablation study of key components}
   \begin{center}
   \resizebox{1\linewidth}{!}{
   \begin{tabular}{c |c |c}
   \toprule[1pt]
   Dataset & Ablation studies of the key components of KaSLA&  EX \\
   \midrule
   \multirow{5}{*}{BIRD-dev} &CodeS + KaSLA & 	60.17  \\
   \cmidrule(lr){2-3}
  & w/o Binary scoring model &   56.06 	\\
  & w/o Probabilistic scoring model &  57.82  	\\
  & w/o Hierarchical strategy & 58.34   	\\
  & w/o upper limit 1 in Eq.~\ref{eq. Relevance scoring estimation} &  53.26  	\\
   \bottomrule[1pt]
   \end{tabular}}
   \end{center}
   \end{table}

\subsubsection{Key Components in KaSLA}
As shown in Table~\ref{tb: ablation study of key components}: (i) Removing the binary scoring model leads to less confirmation for high-confidence elements, causing more redundant ones and highlighting the binary model’s importance in filtering out less relevant elements. (ii) Omitting the probabilistic model risks missing relevant elements, as its soft probabilities effectively compensate for binary scoring errors and offer more robust estimation. (iii) The hierarchical strategy reduces candidate columns; without it, linking must be performed on all columns, adding extra noise and computational cost. (iv) Without the upper limit of 1, the model may introduce disproportionate comparisons, potentially excluding relevant items. This constraint is crucial for balanced relevance estimation and preventing overshadowing of useful elements. These ablation results highlight the significance of KaSLA’s key components.

\begin{figure}[!t]
    \centering
    \subfigure[]{
        \includegraphics[width=0.294\linewidth]{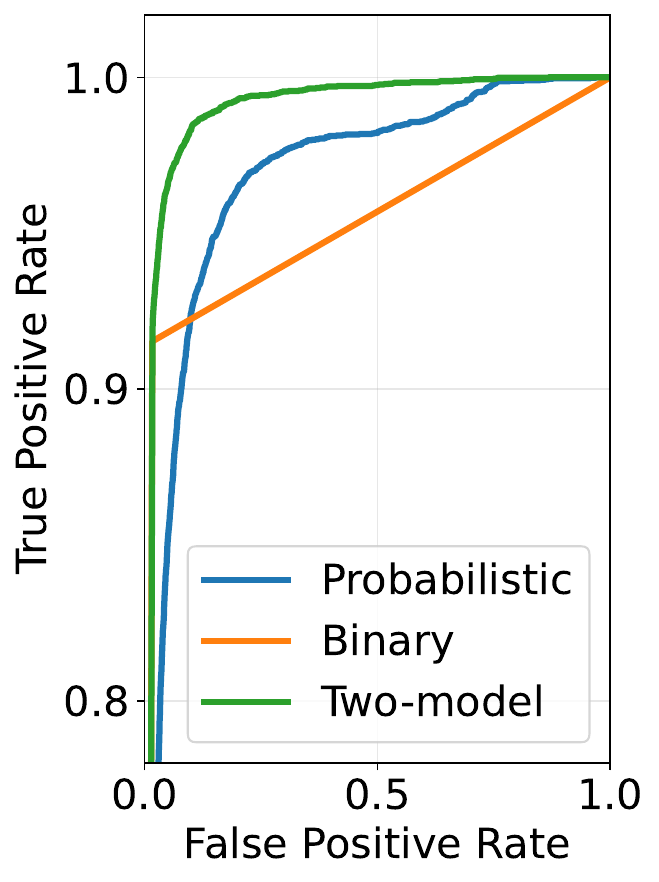}
        \label{fig:ROC_Curves_Comparison}
    }
    \subfigure[]{
        \includegraphics[width=0.29\linewidth]{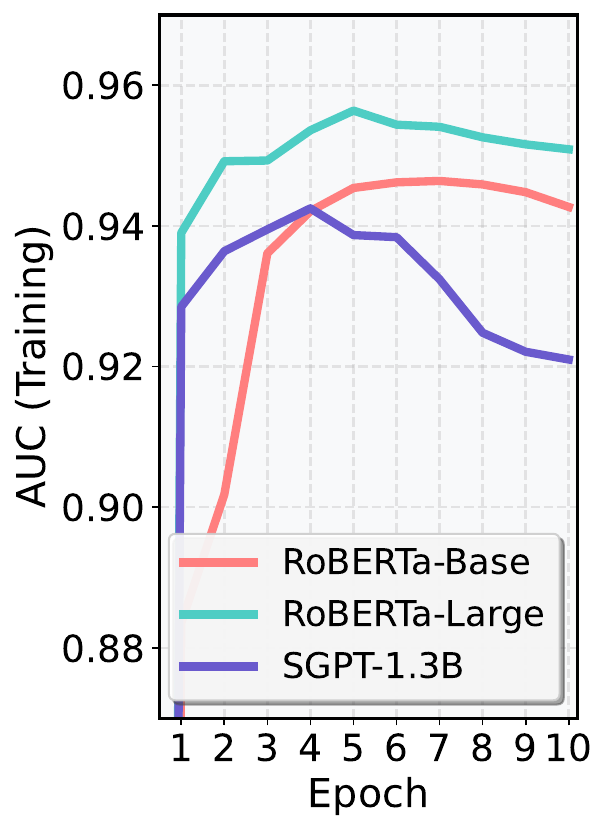}
        \label{fig:ablation_probabilistic_model-training_curves}
    }
    \subfigure[]{
        \includegraphics[width=0.265\linewidth]{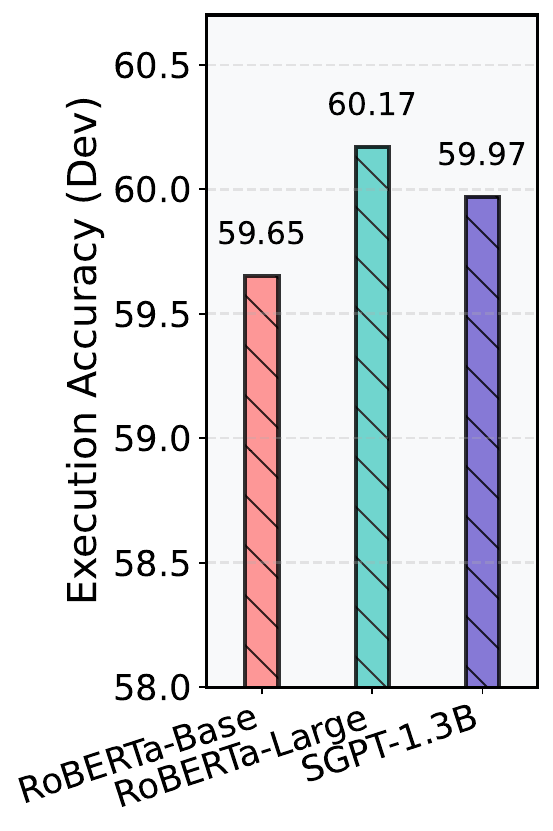}
        \label{fig:ablation_probabilistic_model-accuracy_bar_chart}
    }
    \caption{
        (a) Comparison of ROC curves for the single scoring model and the two-model configuration;
        (b) training accuracy curves of the probabilistic scoring model with different model sizes;
        (c) execution accuracy of CodeS + KaSLA with the probabilistic scoring model for different model sizes.
    }
\end{figure}

To further illustrate the effectiveness of combining the binary and probabilistic models, we present the receiver operating characteristic (ROC) curves in Fig~\ref{fig:ROC_Curves_Comparison}. The results show that the two-model configuration achieves a curve closest to the upper-left corner, indicating better discrimination between relevant and redundant schema elements compared to using either single model alone.

\subsubsection{Probabilistic Scoring Model}
We compare different language models for the probabilistic scoring model in KaSLA, as shown in Fig~\ref{fig:ablation_probabilistic_model-accuracy_bar_chart}. RoBERTa-Large achieves the best performance on BIRD-dev. This demonstrates that RoBERTa-Large’s stronger text representation yields better generalization, making it highly suitable for schema linking tasks. In contrast, RoBERTa-base underfits due to limited capacity, while SGPT-1.3B tends to overfit and fails to improve performance. The training accuracy curves in Fig~\ref{fig:ablation_probabilistic_model-training_curves} further confirm these trends: RoBERTa-Large maintains consistently high and stable accuracy, RoBERTa-base remains lower throughout, and SGPT-1.3B peaks quickly but then drops sharply, indicating severe overfitting. Overall, RoBERTa-Large provides the best balance for the probabilistic scoring model.

\subsubsection{Binary Scoring Model}
\begin{figure}[!t]
    \centering
    \subfigure[]{
        \includegraphics[width=0.42\linewidth]{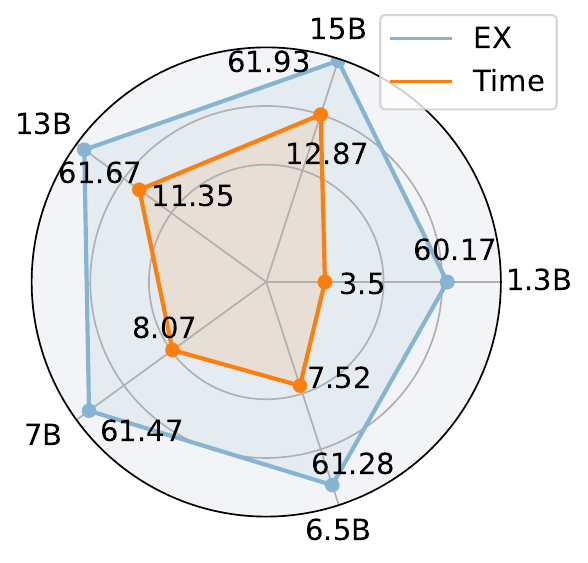}
        \label{fig:ablation_binary_llm_scaling_up_time_cost}
    }
    \subfigure[]{
        \includegraphics[width=0.51\linewidth]{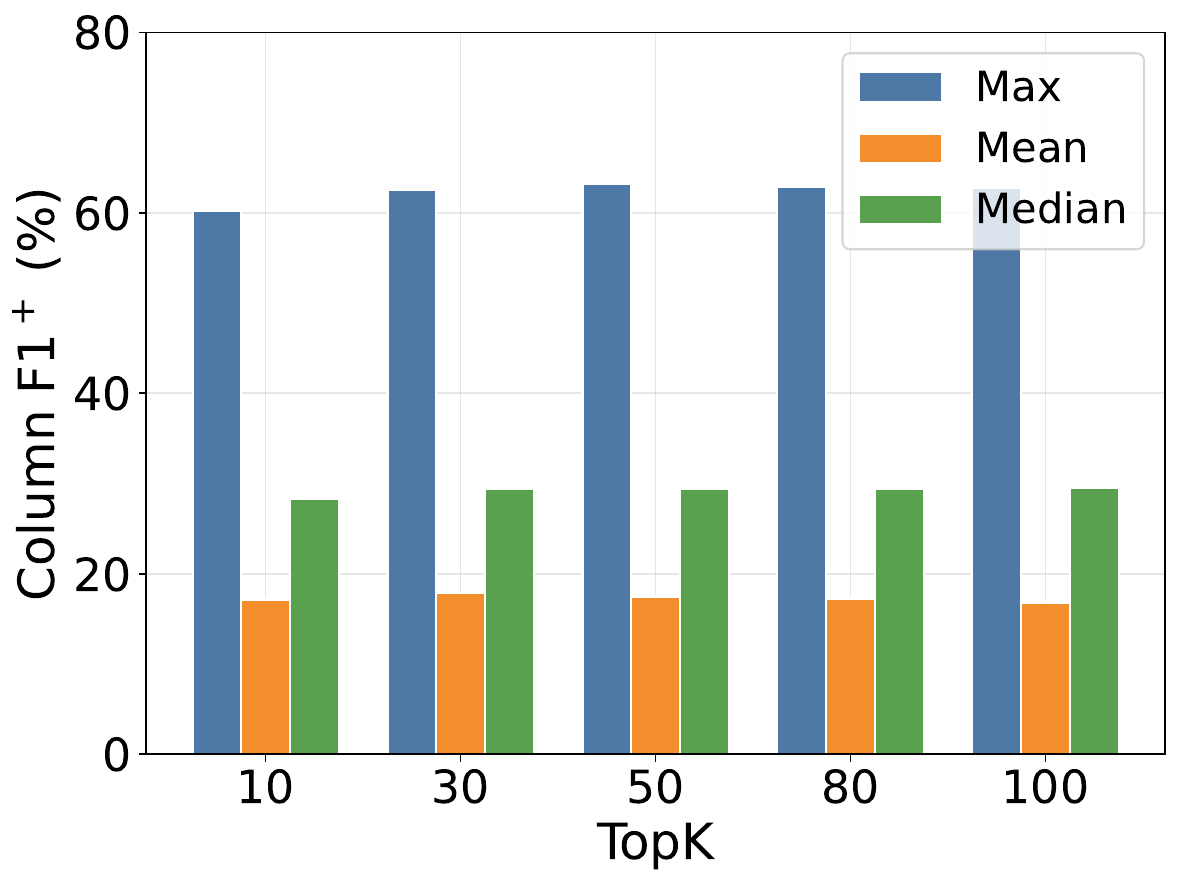}
        \label{fig:topKandCapFunction}
    }
    \caption{
        (a) Effect of model parameter scaling of binary scoring model;
        (b) Ablation study of the redundancy tolerance estimation $f_{\text{tolerance}}$ with different functions and top-K values.
    }
\end{figure}

We present ablation studies on LLM selection with parameter scaling for KaSLA's probabilistic scoring component in Fig~\ref{fig:ablation_binary_llm_scaling_up_time_cost}. We find that although using larger LLMs as the binary scoring model leads to improved performance, the difference in execution accuracy between Deepseek-coder-1.3B and the larger models is marginal. Importantly, Deepseek-coder-1.3B offers clear advantages in terms of faster inference time due to its smaller parameter size. These results indicate that, within the knapsack optimization framework of KaSLA, a low-parameter LLM can achieve performance comparable to that of much larger models. This finding demonstrates that KaSLA is well-suited for efficient and scalable deployment in practical text-to-SQL applications, where smaller models are often preferred due to constraints on computational resources and the need for faster inference.

\begin{table}[t]
   \caption{Performance of CodeS + KaSLA on BIRD-dev with different redundancy estimation functions of $f_{\text{redundancy}}$.}
   \vspace{-4mm}
   \label{tb: ablation study of redundancy estimation function}
   \begin{center}
   \resizebox{1\linewidth}{!}{
   \begin{tabular}{c |c|cc}
   \toprule[1pt]
    Type& For $\left\{r_i \in {\gR}_{\gS_q}\right\}$, $f_{\text{redundancy}}(q, {\gR}_{\gS_q}): $       & Column F1$^+$ &  EX \\
   \midrule
   Inverse& $w_i = {r_i}^{-1} $  & 62.78 & 60.17 \\[2pt]
   Linear$_1$ &$w_i = 1 + (w_{\max} - 1)(1 - r_i), w_{\max}=10$  & 62.39 & 60.04 \\[2pt]
   Linear$_2$ &$w_i = 1 + (w_{\max} - 1)(1 - r_i), w_{\max}=5$  & 61.89 & 59.97 \\[2pt]
  Piece$_1$&   $w_i =\left\{\begin{array}{ll}
     3, & r_i < t_1 \\
     2, & t_1 \leq r_i < t_2 \\
     1, & r_i \geq t_2
   \end{array} \right.,t_1=1/2, t_2=3/4$   & 60.15 & 59.78 \\[2pt]
  Piece$_2$&   $w_i =\left\{\begin{array}{ll}
     3, & r_i < t_1 \\
     2, & t_1 \leq r_i < t_2 \\
     1, & r_i \geq t_2
   \end{array} \right.,t_1=1/3, t_2=2/3$   & 59.84 & 59.58 \\[2pt]
   \bottomrule[1pt]
   \end{tabular}}
   \end{center}
\end{table}

\subsubsection{Redundancy Estimation $f_{\text{redundancy}}$}\label{ablation_Redundancy Estimation}

We conduct an ablation study to systematically evaluate different designs for the redundancy estimation function $f_{\text{redundancy}}$, as summarized in Table~\ref{tb: ablation study of redundancy estimation function}. According to our framework, redundancy should be negatively correlated with relevance: elements with lower relevance scores are more likely to be redundant, and should thus receive higher redundancy weights. To empirically test this principle, we compare several functions with varying degrees of negative correlation between redundancy and relevance: the originally proposed inverse relevance function, two linear mappings with different maximum weights, and two piecewise functions with incrementally weaker negative correlation. As shown in the table, the inverse relevance function, where the negative correlation is strongest, achieves the best overall performance. As the negative correlation in the mapping becomes weaker (from inverse to linear to piecewise), we observe a slight and gradual decrease in both column F1$^+$ and execution accuracy, while the overall framework remains robust and effective across all variants. These results strongly support our central claim that redundancy estimation should be negatively correlated with relevance: the stronger this relationship, the more effectively the redundancy score suppresses unnecessary linking and improves text-to-SQL accuracy.

\subsubsection{Redundancy Tolerance Estimation $f_{\text{tolerance}}$}\label{ablation_Redundancy Tolerance Estimation}

We further conduct ablation studies on the design of the redundancy tolerance function $f_{\text{tolerance}}$, as shown in Fig~\ref{fig:topKandCapFunction}. We compare three options for aggregating redundancy scores from the most similar queries: max, mean, and median. The results demonstrate that using the max function yields significantly higher column F1 scores than mean or median, confirming that a more relaxed upper bound is beneficial. This is because a looser tolerance mitigates the risk of missing relevant elements, while mean or median are too restrictive and lead to poor linking performance.Additionally, we study the effect of varying the top-K value used for retrieving similar queries. The max-based approach in KaSLA achieves stable and robust performance across different K values, indicating insensitivity to this hyperparameter.

\subsection{Empirical Studies (\textbf{RQ4})}
To address \textbf{RQ4}, we conduct empirical studies to evaluate the practical effectiveness of KaSLA. We first assess KaSLA's transferability across domains and scenarios, as summarized in Table~\ref{tb: seen and un-seen data in BIRD-dev.} and Table~\ref{tb: trained on Spider-train but evaluated on BIRD-dev.}. We then evaluate its performance on complex multi-join instances (Table~\ref{tab:performance_comparison on the multi-join instances} and Table~\ref{tab:execution_accuracy_comparison-Spider2.0-lite}), further demonstrating its capability in handling challenging cases. We also present a complexity analysis, including inference time cost (Table~\ref{tb: Inference time cost}), to highlight KaSLA's efficiency. Finally, we provide a systematic failure case analysis on BIRD-dev to offer further insights into real-world performance.

\subsubsection{Transferability} \label{sec: Transferability of KaSLA}
In real-world scenarios, databases come from diverse domains, leading to varied data and queries. Unlike open-source datasets, real-world tasks often lack enough in-domain training data, so cross-domain transferability of schema linking models is essential.
We specifically evaluate KaSLA's performance in cross-domain and cross-scenario settings.

\begin{figure}[!t]
	\centering
        \includegraphics[width=1.0\linewidth]{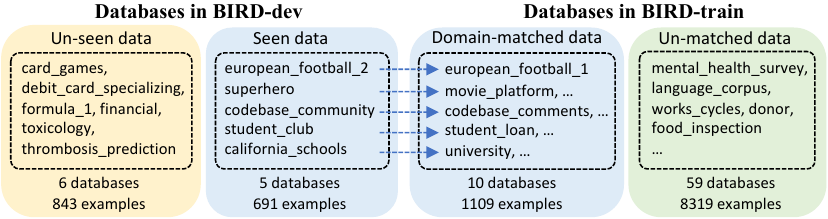}
        \caption{The seen and unseen data in BIRD-dev relative to BIRD-train.} 
        \label{fig:transfer_bird_unseen}
\end{figure}

\begin{table}[!t]
   \caption{Execution Accuracy (EX) (\%) of text-to-SQL models evaluated on seen data and un-seen data in BIRD-dev.}
   \label{tb: seen and un-seen data in BIRD-dev.}
   \begin{center}
   \resizebox{\linewidth}{!}{
   \scalebox{1}{
   \begin{tabular}{c |c c  c }
   \toprule[1pt]
   Evaluation dataset & SuperSQL & CodeS & CodeS + KaSLA \\
   \midrule
   Original BIRD-dev & 58.60  &57.63 &\textbf{60.17 } \\
   Seen data in BIRD-dev & 69.90 &	67.58 	&\textbf{71.06 }  \\
   Un-seen data in BIRD-dev & 49.35 &	51.01 	&\textbf{86.48 }  \\
   \bottomrule[1pt]
   \end{tabular}}}
   \end{center}
   \end{table}

\paragraph{Cross-domain Transferability}
We assess KaSLA’s ability to generalize across database domains using the BIRD dataset, which contains both seen and unseen domains in BIRD-dev relative to BIRD-train. As illustrated in Fig~\ref{fig:transfer_bird_unseen}, seen databases such as ``european\_football\_2'' in BIRD-dev match with ``european\_football\_1'' in BIRD-train, while unseen databases like ``card\_games'' in BIRD-dev do not have any domain-matched database in BIRD-train. Evaluating performance on these unseen domains tests genuine cross-domain transferability. Table~\ref{tb: seen and un-seen data in BIRD-dev.} shows that KaSLA consistently outperforms baseline models on both seen and unseen domains. This strong performance is attributed to the robustness of the KaSLA optimization framework, which uses knapsack optimization to reduce missing and redundant elements. It enables KaSLA to achieve optimal results even on unseen domains after training on open-source datasets.

\paragraph{Cross-scenario Transferability}
To further evaluate KaSLA’s transferability, we test its cross-scenario performance between different datasets. We train the scoring model on BIRD-train and test on Spider-dev, and vice versa. In addition, we further investigate the impact of transferring the query-linking pool across datasets. Specifically, we use the query-linking pool constructed from one dataset (e.g., Spider-train) while applying KaSLA to another dataset (e.g., BIRD-dev).

Results in Table~\ref{tb: trained on Spider-train but evaluated on BIRD-dev.} show that KaSLA remains robust even when the query-linking pool is transferred from a different dataset, achieving performance very close to using a pool constructed from the same domain. Models trained on a different dataset still outperform baselines and are only slightly worse than their domain-matched counterparts. These results demonstrate that KaSLA has strong cross-scenario transferability, highlighting its robustness when applied to new text-to-SQL scenarios. This also confirms that KaSLA, when trained on open-source datasets or with a query-linking pool from another scenario, can be directly and effectively applied to unseen scenarios in real-world applications.

\begin{figure}[!t]
	\centering
        \includegraphics[width=1.0\linewidth]{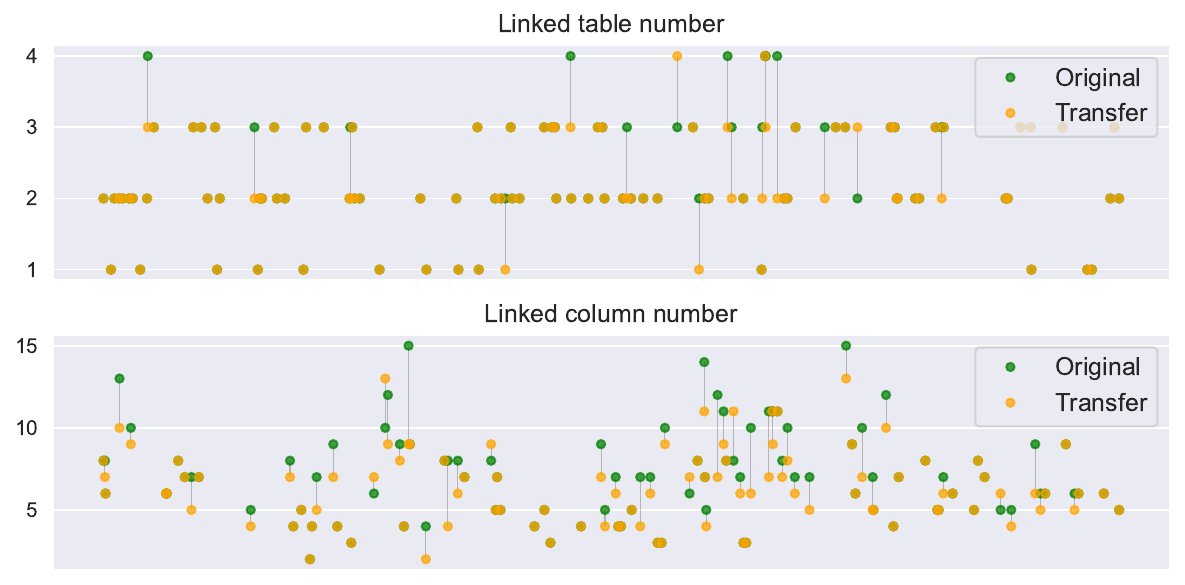}
        \caption{Comparison of the number of linked tables and columns using the original and transfer query-linking pools in KaSLA. Each pair of points represents the results for the same instance under the two settings.}
        \label{fig:Transfer_Query_Linking_Pool_Paired-Comparison}
\end{figure}

\begin{table}[!t]
   \caption{Performance of KaSLA trained on Spider-train but evaluated on BIRD-dev, and vice versa, for cross-scenario transfer ability evaluation.}
   \vspace{-4mm}
   \label{tb: trained on Spider-train but evaluated on BIRD-dev.}
   \begin{center}
   \resizebox{\linewidth}{!}{
   \scalebox{1}{
   \begin{tabular}{c | c |c | c c c}
   \toprule[1pt]
   \multirow{2}{*}{Method} & \multicolumn{2}{c|}{Transfer from Spider-train} & \multicolumn{3}{c}{BIRD-dev}   \\
   \cmidrule(lr){2-3} \cmidrule(lr){4-6}
    & Scoring Models& Pool &Table F1$^+$&Column F1$^+$ & EX \\
   \midrule
   CodeS & -&- &58.21  &29.56   &57.63  \\
   \midrule
   \multirow{4}{*}{CodeS + KaSLA} 
   & -&-  &82.21   &62.53    & 60.17  \\
    & \checkmark &- &81.09  & 60.23 & 60.02 \\
    & - &\checkmark  &81.67&60.65  &60.10   \\
    & \checkmark &\checkmark  &80.72  &59.82   &59.97   \\
   \bottomrule[1pt]
   \end{tabular}}}
   \end{center}
   \end{table}

To further illustrate the effect of transferring the query-linking pool, Fig~\ref{fig:Transfer_Query_Linking_Pool_Paired-Comparison} visualizes the number of linked tables and columns for each instance under both the original and transfer settings. The paired dot/line plots show that table linking is highly consistent across both settings, while a slight deviation is observed in column linking when using the transferred pool. This minor difference explains the small drop in performance seen in Table~\ref{tb: trained on Spider-train but evaluated on BIRD-dev.}. Future work could focus on enhancing the transferability of query-linking pool designs to further improve cross-scenario robustness. Additionally, we test KaSLA on Spider2.0-lite~\cite{lei2024spider}, a dataset without a training set, where both the scoring model and the query-linking pool are transferred from BIRD. As shown in Table~\ref{tab:execution_accuracy_comparison-Spider2.0-lite}, KaSLA brings significant improvements to both E-SQL and RSL-SQL models. These results highlight KaSLA’s strong robustness and effectiveness when applied directly to new, unseen scenarios.

\subsubsection{Performance on Multi-join Instances}

To assess KaSLA’s effectiveness on complex schema linking scenarios, we evaluate its performance on BIRD-dev by grouping instances according to the number of tables needed in each instance, as shown in Table~\ref{tab:performance_comparison on the multi-join instances}. Across all groups, including single-table, two-table, and multi-join (three or more tables) queries, KaSLA consistently achieves substantial improvements over the baseline. Notably, the largest relative gains are observed in the most complex multi-join cases. This indicates that KaSLA’s optimization framework can robustly handle challenging multi-hop join scenarios. However, a gap remains for multi-joins compared with single- and two-table cases, suggesting that further refinement in bridging long join paths could provide additional benefits in future work. Further, as shown in Table~\ref{tab:execution_accuracy_comparison-Spider2.0-lite}, KaSLA also delivers stable improvements on the complex and cross-domain Spider2.0-lite dataset. This result further confirms KaSLA's robustness and effectiveness in handling challenging multi-join queries and complex schema structures.

\begin{table}[!t]
\centering
\caption{Performance comparison of CodeS and CodeS+KaSLA on the multi-join instances.}
\label{tab:performance_comparison on the multi-join instances}

\resizebox{\linewidth}{!}{
\begin{tabular}{cccccc}
\toprule
\multirow{2}{*}{Table} & \multirow{2}{*}{Instances} & \multicolumn{2}{c}{Table F1$^+$} & \multicolumn{2}{c}{EX} \\
\cmidrule(lr){3-4} \cmidrule(lr){5-6}
 & & CodeS &CodeS + KaSLA & CodeS & CodeS + KaSLA \\
\midrule
1 & 362 & 38.67 & 73.61 & 60.22 & 64.92 \colorbox{gray!20}{\scriptsize $\uparrow$ 7.80\%}\\
2 & 942  & 64.05 & 86.02& 58.60 & 60.08 \colorbox{gray!20}{\scriptsize $\uparrow$ 2.53\%}\\
$\geq$ 3 & 230  &  73.72 & 81.36& 48.26 & 52.17 \colorbox{gray!20}{\scriptsize $\uparrow$ 8.10\%}\\
\bottomrule
\end{tabular}}
\end{table}

\begin{table}[!t]
\centering
\caption{Execution Accuracy (EX) of KaSLA on Spider2.0-lite. We use DeepSeek-V3 as the SQL generation model.}
\label{tab:execution_accuracy_comparison-Spider2.0-lite}

\resizebox{\linewidth}{!}{
\begin{tabular}{ccc | cc }
\toprule
Instances & E-SQL & E-SQL + KaSLA & RSL-SQL & RSL-SQL + KaSLA \\
\midrule
Random 100 examples & 21.00\% & 24.00\% \colorbox{gray!20}{\scriptsize $\uparrow$ 14.29\%}& 25.00\% & 28.00\% \colorbox{gray!20}{\scriptsize $\uparrow$ 12.00\%}\\
\bottomrule
\end{tabular}}
\end{table}

   \begin{table}[!t]
   \caption{Inference time cost per instance of each component in KaSLA using CodeS-15B as the text-to-SQL model.}
   \label{tb: Inference time cost}
   \begin{center}
   \scalebox{1}{
   \begin{tabular}{l c c}
   \toprule
   Model & Full Schema & KaSLA \\
   \midrule
   Binary scoring function & / & 3.5 s \\
   Probabilistic scoring model& / & 0.12 s \\
   Estimation and dynamic programming & / & $<$ 0.01 s \\
   SQL generation with CodeS-15B & 5.15 s & 1.85 s \\
   \midrule
   Total time & 5.15 s & 5.48 s \\
   \midrule
   EX & 57.63 & 60.17 \\
   \bottomrule
   \end{tabular}}
   \end{center}
   \end{table}

\subsubsection{Complexity Analysis and Inference Time Cost}\label{Complexity Analysis and Inference Time Cost}
We provide a complexity analysis for offering insights into KaSLA's implementation and computational efficiency. The efficiency of KaSLA stems from its hierarchical approach, dynamic programming and inherent parallelism. KaSLA's time complexity is determined by two main phases: table linking and column linking within the linked tables. For a database with \( |\gT| \) tables and a table tolerance of \( u^{\gT} \), the table linking phase has a time complexity of \( O(|\gT| u^{\gT}) \). KaSLA then links columns within the selected table. Let \( |\gC^t| \) and \( u^{\gC} \) represent the number of columns and the column capacity of a selected table, respectively. The time complexity of the column linking phase is \( O(|\gC^t| u^{\gC}) \). Consequently, the total time complexity of KaSLA is \( O(|\gT| u^{\gT} + |\gT|(|\gC^t| u^{\gC})) \). Since all factors are constants, this results in linear complexity.
 
We evaluated the inference time cost of KaSLA in Table~\ref{tb: Inference time cost}. Although KaSLA introduces lightweight steps such as binary/probabilistic scoring (totaling 3.62s), the SQL generation phase is significantly accelerated (1.85s vs. 5.15s) due to prompt reduction from precise schema linking. The overall per-instance inference time only increases marginally (5.48s vs. 5.15s), while executable accuracy improves (60.17 vs. 57.63). This demonstrates that KaSLA provides superior accuracy and efficient computation by minimizing redundant schema elements with negligible additional cost.

\subsubsection{Failure Case Analysis}\label{section: Failure case analysis}

\begin{figure}[!t]
	\centering
        \includegraphics[width=1.0\linewidth]{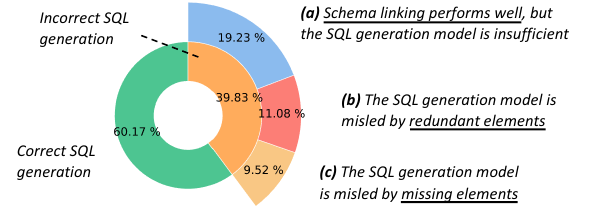}
        \caption{Failure case analysis of ``CodeS + KaSLA''  in BRD-dev.} 
        \label{fig:sql_pie_charts}
\end{figure}

We conduct a comprehensive failure case analysis of “CodeS + KaSLA” on BIRD-dev, as illustrated in Fig~\ref{fig:sql_pie_charts}. The results show that KaSLA’s schema linking is effective in the vast majority of cases, though a portion of correct linkings still result in incorrect SQL due to limitations in the generation model. Misleading redundant or missing elements within the schema linking stage directly account for the remaining errors in SQL prediction. These findings highlight both the effectiveness of KaSLA in minimizing schema linking errors and the importance of continued improvement, either by refining the schema linking optimization or by leveraging stronger SQL generation models.

\section{Related Work}
\label{sec:Related_work}

\subsection{LLM-based Text-to-SQL}
Large language models (LLMs) have propelled two primary approaches for text-to-SQL: in-context learning (ICL) and fine-tuning. ICL (prompt engineering)~\cite{reynolds2021prompt} has become central, with early efforts analyzing prompt designs and LLM architectures~\cite{liu2023comprehensive}. Effective ICL strategies adopted from other NLP tasks, like Chain-of-Thought (CoT) prompting~\cite{wei2022chain, wei2021finetuned}, have advanced text-to-SQL generation~\cite{tai2023exploring,li2025prompt4vis}. Methods such as DIN-SQL~\cite{pourreza2023dinsql}, ACT-SQL~\cite{zhang2023act}, and DAIL-SQL~\cite{gao2023dailsql} enhance CoT with additional examples or knowledge~\cite{hong2024knowledge}, while others employ multi-agent refinement~\cite{wang2024macsql, talaei2024chess} or online example generation~\cite{pourreza2024chase}. As a result, ICL-based models such as SuperSQL~\cite{li2024dawn}, Dubo-SQL~\cite{thorpe2024dubo}, TA-SQL~\cite{qu2024before}, E-SQL~\cite{caferouglu2024sql}, CHESS~\cite{talaei2024chess}, and RSL-SQL~\cite{cao2024rsl} define the state of the art for LLM-based text-to-SQL~\cite{hong2024next}. Fine-tuning is the other mainstream strategy~\cite{wei2021finetuned}, especially useful for complex queries and real-world databases~\cite{ li2023BIRD,lei2024spider}. Fine-tuning \cite{wei2021finetuned, gao2023dailsql} is applied to enhance open-source LLMs for SQL generation. Advanced methodologies include the two-stage framework introduced in CodeS \cite{li2024codes}, logic-guided demonstration retrieval in PURPLE \cite{ren2024purple}, dual-model schema and SQL generation in DTS-SQL \cite{pourreza2024dtssql}, ensemble refinement in XiYan-SQL \cite{gao2024xiyan}, and reinforcement learning employed in SQL-R1 \cite{ma2025sql}. Collectively, these ICL and fine-tuning techniques are steadily and significantly advancing the frontier of LLM-based text-to-SQL.

\subsection{Schema Linking in Text-to-SQL}
Within the text-to-SQL task, schema linking is an essential module for aligning natural language queries with database contents, playing a pivotal role in SQL generation~\cite{hong2024next}. Recent works employ LLMs to enhance schema linkage via semantic retrieval, historical query integration, and bidirectional linking~\cite{pourreza2024dtssql, cao2024rsl, zhang2024structure}. Methods like DAIL-SQL~\cite{gao2023dailsql} and E-SQL~\cite{caferouglu2024sql} exploit LLMs' contextual understanding for linkage refinement~\cite{gao2023dailsql, caferouglu2024sql}, while retrieval-based strategies leverage database information to improve SQL generation~\cite{pourreza2024dtssql}. CodeS~\cite{li2024codes} utilizes semantic retrieval for better schema linking. CHESS~\cite{talaei2024chess} introduces schema selector agents that prompt LLMs to select relevant tables and columns. Yet, many state-of-the-art text-to-SQL models focus on fine-tuning strategies. DTS-SQL~\cite{pourreza2024dtssql} fine-tunes a local LLM for table selection, and Solid-SQL~\cite{liu-etal-2025-solid} uses data augmentation to boost schema linking. Other approaches, such as TA-SQL~\cite{qu2024before}, employ in-context learning or, like RSL-SQL~\cite{cao2024rsl}, bidirectional simplification. Despite these efforts, E-SQL~\cite{qu2024before} and Distillery-SQL~\cite{maamari2024death} point out that schema linking challenges persist, prompting some models to input the entire schema directly for SQL generation.

Different with the previous research, our KaSLA utilize a knapsack optimization framework to avoid the relevant elements missing and the inclusion of redundant elements, optimizing schema linking to enhance SQL generation accuracy effectively.

\section{Limitation}
\label{sec:Limitation}
Although KaSLA consistently improves schema linking and SQL generation performance across benchmarks, there are still minor limitations that point to meaningful future research directions. First, our current criterion for redundancy is related to relevance. While this formulation has proven effective in our experiments, exploring independent modeling of redundancy and relevance within the knapsack framework could be a promising direction. Such decoupling may further advance knapsack-based schema linking in text-to-SQL. Second, although we show that transferring the scoring models and query-linking pool from other domains yields results comparable to using same-domain resources, further enhancing KaSLA’s transferability to achieve equivalent performance across domains remains a valuable research direction.

\section{Conclusion}
\label{sec:Conclusion}
This paper studies schema linking in text-to-SQL, focusing on linking relevant elements while avoiding missing and redundant ones. We propose Recall$^+$ and Precision$^+$ with a restricted missing-element indicator for better linking assessment. Furthermore, we introduce the Knapsack Schema Linking Approach (KaSLA), a method for overcoming schema linking challenges in text-to-SQL tasks. Based on knapsack optimization, KaSLA links relevant elements while preventing missing and redundant ones, significantly improving SQL generation accuracy. KaSLA uses a hierarchical strategy, first linking tables, then columns within selected tables to reduce search space. In each step, KaSLA applies knapsack optimization with limited redundancy tolerance. Experiments on Spider and BIRD benchmarks show KaSLA greatly enhances SQL generation by replacing existing schema linking. These results highlight KaSLA's potential to advance schema linking and text-to-SQL capabilities, offering a robust solution for more precise, efficient database interactions.

\section*{Acknowledgement}
\label{sec:Acknowledgement}
The work described in this paper was fully supported by a grant from the Innovation and Technology Commission of the Hong Kong Special Administrative Region, China (Project No. GHP/391/22).

\section*{AI-Generated Content Acknowledgement}
\label{sec:AI-gen Acknowledgement}
LLMs were employed solely for minor editorial assistance, such as correcting typos and refining grammar. All conceptual development, algorithmic design, and experimental work were performed independently by the authors without using LLMs.

\nocite{*}
\bibliographystyle{IEEEtran}

\bibliography{reference}

\end{document}